\pdfoutput=1
\documentclass[11pt]{article}

\usepackage{acl}

\usepackage{times}
\usepackage{latexsym}
\usepackage{graphicx}
\usepackage{subfigure}
\usepackage{multirow}
\usepackage{multicol}
\usepackage{booktabs}
\usepackage{amssymb}
\usepackage{bbding}
\usepackage{amsmath}
\usepackage{amsfonts}
\usepackage{amssymb}
\usepackage{hyperref}
\usepackage{threeparttable}
\usepackage{pifont}

\usepackage[T1]{fontenc}
\usepackage[utf8]{inputenc}

\usepackage{microtype}

\title{Discrete Cross-Modal Alignment Enables Zero-Shot Speech Translation}

\author{Chen Wang\textsuperscript{1}\thanks{\ \ Work was done while at Alibaba DAMO Academy.}, Yuchen Liu\textsuperscript{1}, Boxing Chen\textsuperscript{2}, Jiajun Zhang\textsuperscript{1}\thanks{\ \ Corresponding author.}, \\
{\bf Wei Luo\textsuperscript{2}, Zhongqiang Huang\textsuperscript{2}, Chengqing Zong\textsuperscript{1}} \\
\textsuperscript{1} National Laboratory of Pattern Recognition, Institute of Automation, CAS, Beijing, China \\
\textsuperscript{2} Machine Intelligence Technology Lab, Alibaba DAMO Academy\\
\texttt{$\left\{\right.$wangchen2020,yuchen.liu$\left .\right\}$@ia.ac.cn}, \texttt{$\left\{\right.$jjzhang,cqzong$\left .\right\}$@nlpr.ia.ac.cn} \\
\texttt{$\left\{\right.$Boxing.cbx,muzhuo.lw,z.huang$\left .\right\}$@alibaba-inc.com}
}

\begin{document}
\maketitle
\begin{abstract}

End-to-end Speech Translation (ST) aims at translating the source language speech into target language text without generating the intermediate transcriptions.
However, the training of end-to-end methods relies on parallel ST data, which are difficult and expensive to obtain. 
Fortunately, the supervised data for automatic speech recognition (ASR) and machine translation (MT) are usually more accessible, making zero-shot speech translation a potential direction.
Existing zero-shot methods fail to align the two modalities of speech and text into a shared semantic space, resulting in much worse performance compared to the supervised ST methods.
In order to enable zero-shot ST, we propose a novel \textbf{D}iscrete \textbf{C}ross-\textbf{M}odal \textbf{A}lignment (DCMA) method that employs a shared discrete vocabulary space to accommodate and match both modalities of speech and text. Specifically, we introduce a vector quantization module to discretize the continuous representations of speech and text into a finite set of virtual tokens, and use ASR data to map corresponding speech and text to the same virtual token in a shared codebook. This way, source language speech can be embedded in the same semantic space as the source language text, which can be then transformed into target language text with an MT module.
Experiments on multiple language pairs demonstrate that our zero-shot ST method significantly improves the SOTA, and even performs on par with the strong supervised ST baselines\footnote{Our code is available at \url{https://github.com/ZNLP/zero-shot-st}}.

\end{abstract}

\section{Introduction}

End-to-end Speech Translation (ST) aims at designing a single model to directly learn the mapping between source language speech and target language text, and has attracted much attention recently due to its advantages of no error propagation and lower decoding latency
\cite{DBLP:journals/corr/BerardPSB16,DBLP:conf/interspeech/LiuXZHWWZ19,DBLP:conf/aaai/WangWLY020,DBLP:conf/acl/XuHLZHJXZ20}. However, the training of end-to-end models requires large-scale and high-quality parallel ST data, which are expensive and difficult to obtain. Public datasets such as MUST-C \cite{DBLP:conf/naacl/GangiCBNT19} and CoVoST \cite{DBLP:conf/lrec/WangPWG20} are quite limited in scale and languages. 
In contrast, the datasets for automatic speech recognition (ASR) and machine translation (MT) are easier to access in practice. Therefore, zero-shot ST, which learns an end-to-end model using only ASR and MT data, is a direction worth exploring.

\begin{figure}[t]
    \centering
    \centering
    \includegraphics[width=\linewidth]{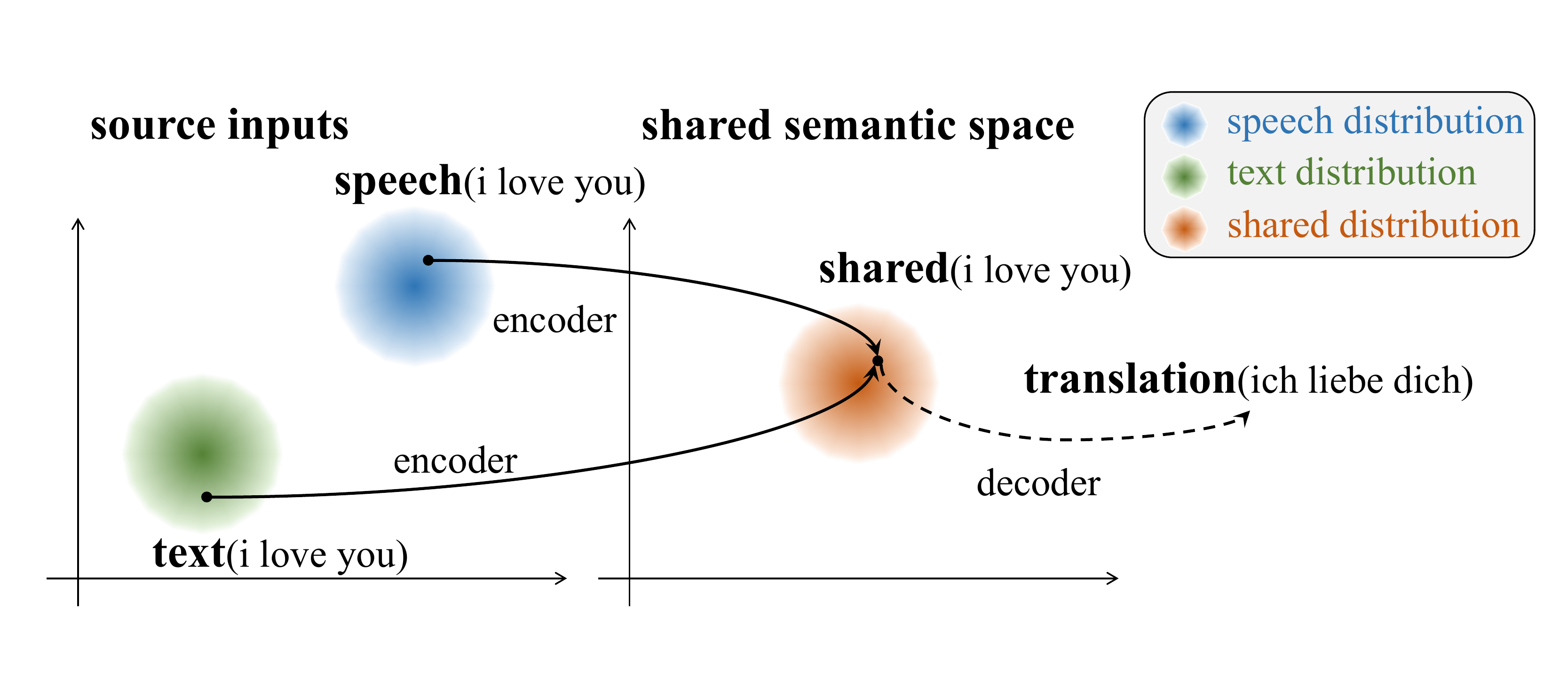}
    \caption{The key of zero-shot speech translation is to learn an appropriate shared semantic space for source language speech and text.}
    \label{fig:intro}
\end{figure}

As illustrated in Figure \ref{fig:intro}, the key of zero-shot ST is to learn an appropriate shared semantic space for source language speech and text, after which the model can translate from the common space using an MT module without relying any ST data. There are some attempts to achieve zero-shot ST using multi-task learning that implicitly aligns speech and text \cite{DBLP:conf/asru/EscolanoCFS21,DBLP:journals/corr/abs-2107-06010}. Due to the lack of supervised objective functions for cross-modal alignment, the speech and text representations cannot be well aligned in these methods and the results lag behind supervised settings by a significant margin.
Inspired by recent research on learning shared semantic space for speech and text \citep{DBLP:conf/emnlp/AlinejadS20,DBLP:journals/corr/abs-2010-14920,DBLP:conf/acl/HanWJL21}, we aim to design a supervised cross-modal alignment task that explicitly maps speech and text into a common feature space. 

However, our preliminary study indicates that the existing cross-modal alignment methods that bridge the gap between speech and text in a continuous space do not work well in the zero-shot ST task because in this condition, it is difficult to completely align two modalities to the same distribution in a high-dimensional continuous space. To address this issue, we propose a novel \textbf{D}iscrete \textbf{C}ross-\textbf{M}odal \textbf{A}lignment (DCMA) method that employs a shared discrete vocabulary space to accommodate and match both modalities of speech and text. With the shared discrete vocabulary across the two modalities, the source language speech and the corresponding text are mapped to the same virtual token, ensuring representational consistency.

Specifically, the ST model consists of a speech encoder and a text decoder. We introduce a vector quantization module between the speech encoder and the text decoder to discretize the continuous representations into a finite set of virtual tokens. The ASR data are used to provide supervision that maps the speech and its corresponding text into the same virtual token of the shared codebook.
In addition to the vector quantization module, the speech encoder is decoupled into an acoustic encoder and a semantic encoder. A shared memory module following the speech encoder is introduced to project variable-length input features of both source language speech and text into fix-sized ones. Machine translation is jointly trained to learn the mapping between the fix-sized features on the source side and the text on the target side. To further enhance the speech encoder, masked language model (MLM) and connectionist temporal classification (CTC) are employed as auxiliary tasks in which all parameters are shared. Experimental results on the benchmark dataset MUST-C demonstrate that our discrete alignment method can significantly improve the performance of zero-shot speech translation.

The contributions of this paper are as follows:
\begin{itemize}
    \item We propose a novel cross-modal alignment method, DCMA, which aligns speech and text in a shared discrete semantic space.
    \item We design a vector quantization module to discretize continuous representations to a finite set of virtual tokens so that cross-modal alignment in discrete space can be well achieved.
    \item Experimental results demonstrate that our method significantly improves the SOTA in zero-shot ST and  performs on par with the supervised models.
\end{itemize}

\section{Related Work}

\paragraph{Data Scarcity in End-to-End ST} 
 
\citet{DBLP:journals/corr/BerardPSB16,DBLP:conf/naacl/DuongACBC16} give the first proof of potential for end-to-end ST models, which have become popular recently \citep{DBLP:conf/acl/InagumaKDKYHW20,DBLP:conf/ijcnlp/WangTMWOP20}. 
However, the performance of the end-to-end methods is heavily dependent on large-scale and high quality parallel data, which are difficult to collect on a large scale \cite{DBLP:conf/naacl/GangiCBNT19,DBLP:conf/lrec/WangPWG20}. Many techniques, such as pretraining \citep{DBLP:conf/icassp/BerardBKP18,DBLP:conf/interspeech/BansalKLLG18,DBLP:conf/naacl/BansalKLLG19,DBLP:conf/acl/WangWLZY20,DBLP:conf/icml/ZhengCM021}, multi-task learning \citep{DBLP:conf/acl/ChuangSLL20,DBLP:conf/acl/XuHLZHJXZ20,DBLP:conf/acl/TangPLWG20,DBLP:conf/icassp/TangPWMG21}, knowledge distillation \citep{DBLP:conf/interspeech/LiuXZHWWZ19,DBLP:conf/clic-it/GaidoGNT20,DBLP:conf/naacl/InagumaKW21}, multilingual translation \citep{DBLP:conf/asru/InagumaDKW19,DBLP:conf/asru/GangiNT19,DBLP:conf/acl/LePWGSB20}, and data augmentation \citep{DBLP:conf/icassp/JiaJMWCCALW19,DBLP:conf/iwslt/BaharZSN19,DBLP:conf/interspeech/WangWPBAC21} are applied to utilize the data from related tasks.  
Zero-shot scenario has attracted attention in recent years, but there still remains a significant performance gap between zero-shot and supervised methods \cite{DBLP:conf/asru/EscolanoCFS21, DBLP:journals/corr/abs-2107-06010}. One contributing factor is the lack of explicit cross-modal alignments. 

\begin{figure*}[t]
    \centering
    \includegraphics[width=0.95\textwidth]{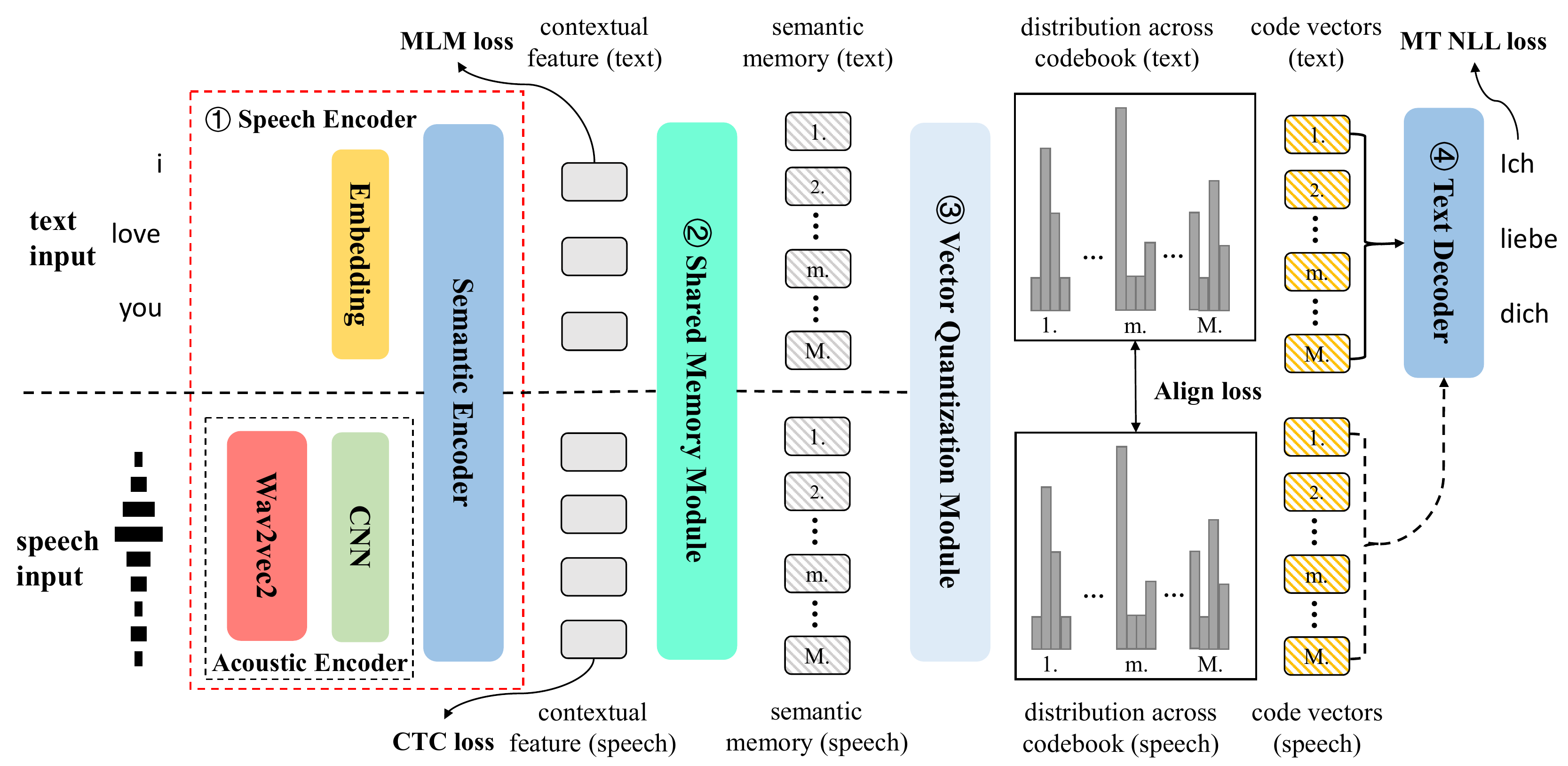}
    \caption{Overview of our model DCMA. The speech encoder is decoupled into acoustic encoder and semantic encoder. We adopt a shared memory module to project feature sequence into a fixed length and a vector quantization module to perform cross-modal alignment in discrete space. The text decoder is used to translate from the discrete common space output by vector quantization module. The part above the dashed line illustrates the data flow of the MT task from source language text into target language text. The ASR data is employed to design the cross-modal alignment loss. The data flow of the bottom part can be used to perform the ST task.
    }
    \label{fig:vq_memory}
\end{figure*}

\paragraph{Cross-modal Alignment in End-to-End ST} 
Cross-modal alignment aims at aligning representations of speech and text to extract common features.
Some recent works have pointed out that the representation gap between speech and text is a major obstacle to speech translation. There are many proposals to bridge the gap, including introducing an alignment task \citep{DBLP:conf/emnlp/AlinejadS20,DBLP:journals/corr/abs-2010-14920,DBLP:conf/acl/TangPLWG20,DBLP:conf/acl/HanWJL21}, mixup strategy \citep{DBLP:journals/corr/abs-2203-10426} and multimodal pretraining \citep{DBLP:conf/icml/ZhengCM021,DBLP:journals/corr/abs-2110-10329,DBLP:journals/corr/abs-2110-07205,DBLP:journals/corr/abs-2111-09296,DBLP:journals/corr/abs-2202-01374}. Additionally, some more sophisticated modules such as adaptive feature selection \citep{DBLP:conf/emnlp/ZhangTHS20}, shrink mechanism \citep{DBLP:journals/corr/abs-2010-14920} and shared memory module \citep{DBLP:conf/acl/HanWJL21} have been proposed to address the length inconsistency problem. These methods align the representations of two modalities in continuous feature space.  
Although these methods work effectively in supervised settings, our preliminary study indicates that the alignments in continuous space do not work well under zero-shot scenario because it is difficult to match two continuous distributions without strong supervision.

\section{Method}

\subsection{Problem Definition}
We attempt to train an end-to-end model with only ASR and MT corpora, achieving zero-shot speech translation. 
We denote the ASR corpus and the MT corpus as $\mathcal{D}_{ASR}=\{(\mathbf{s},\mathbf{x})\}$ and $\mathcal{D}_{MT}=\{(\mathbf{x}^{\prime},\mathbf{y})\}$ respectively, where $\mathbf{s}$ is the audio wave sequence, $\mathbf{x}$ is the corresponding transcripts, $\mathbf{x}^{\prime}$ is the source language text and $\mathbf{y}$ is the corresponding translation in the target language. 

\subsection{Model Architecture}

Our DCMA model follows the encoder-decoder framework, as shown in Figure \ref{fig:vq_memory}. In addition to the conventional speech encoder and text decoder, we also introduce a shared memory module and a shared vector quantization module between the encoder and the decoder.

\paragraph{Speech Encoder}

The speech encoder consists of an acoustic encoder and a semantic encoder to encourage information sharing between tasks \cite{DBLP:conf/aaai/WangWLY020,DBLP:conf/acl/TangPLWG20,DBLP:conf/acl/XuHLZHJXZ20}. For speech input, we use the pretrained wav2vec2.0 \cite{DBLP:conf/nips/BaevskiZMA20} as the acoustic encoder to extract speech representations from the original waveform, which has been shown effective in supervised ST \cite{DBLP:conf/interspeech/YeW021,DBLP:journals/corr/abs-2203-10426}. Since the speech feature sequence can be very long, we add two additional one-dimensional convolution layers with stride 2 to shrink the length by a factor of 4. The speech representations are then fed into the shared semantic encoder to obtain the semantic representations. For text input, only the shared semantic encoder is employed. The semantic encoder follows Transformer \cite{DBLP:conf/nips/VaswaniSPUJGKP17}, and its output is denoted as $\mathbf{H}\in \mathbb{R}^{l\times d}$. 

In order to enhance the semantic encoder so that it can embed both acoustic and textual features, we apply the Connectionist Temporal Classification (CTC) \citep{DBLP:conf/icml/GravesFGS06} on the contextual features of speech and the Masked Language Model (MLM) \citep{DBLP:conf/naacl/DevlinCLT19} on the contextual features of text. The softmax vocabulary and paramters are shared across the two tasks to encourage implicit alignment between the speech and text representations learnt by the semantic encoder \cite{DBLP:journals/corr/abs-2202-01374}. 
Specifically, the text in source language from $\mathcal{D}_{MT}$ follows the same mask policy as BERT \citep{DBLP:conf/naacl/DevlinCLT19}, and the model is required to predict the correct masked tokens. The CTC loss is applied on the speech input from $\mathcal{D}_{ASR}$, using token-level transcription as the target.

\begin{figure}[t]
    \centering
    \includegraphics[width=\linewidth]{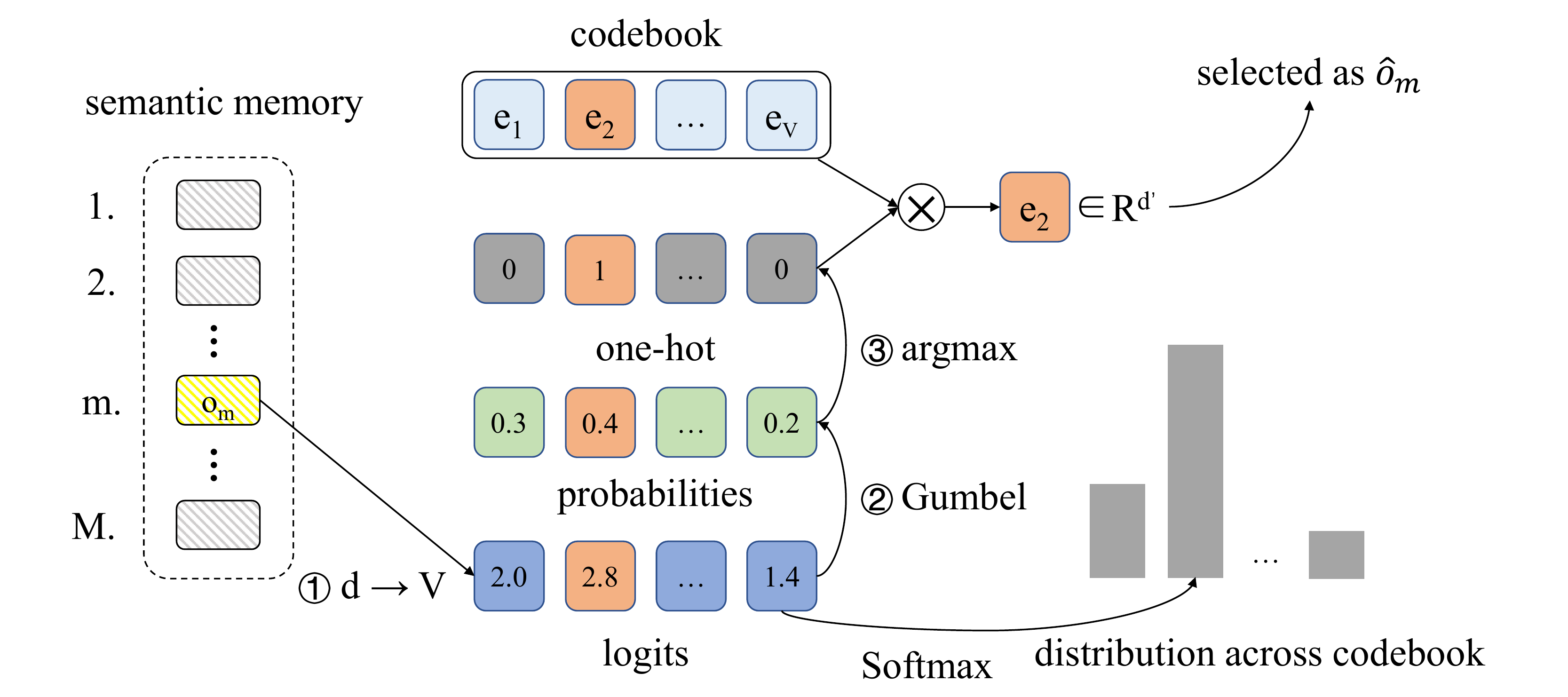}
    \caption{The detailed calculation process of the vector quantization module in one group. This process is performed $G$ times in parallel, and the selected code vectors from $G$ groups are concatenated to get the final output.}
    \label{fig:gumbel}
\end{figure}

\paragraph{Shared Memory Module}

Since the semantic encoder outputs representations in different lengths for speech and text, making it difficult to perform cross-modal alignment, we introduce a shared memory module \cite{DBLP:conf/acl/HanWJL21} to project the contextual features from two modalities with different lengths into fix-sized $M$. In calculating the attention, this module keeps $M$ learnable, modality-independent memory queries, while the contextual features are used as keys and values as shown below:
\begin{gather}
    \mathbf{Q} = \mathbf{M} \in \mathbb{R}^{M\times d} \\
    \mathbf{K} = \mathbf{V} = \mathbf{H} \in \mathbb{R}^{l\times d} \\
    \mathbf{O} = \text{MultiHead}(\mathbf{Q},\mathbf{K},\mathbf{V}) \in \mathbb{R}^{M\times d}
\end{gather}
where $\mathbf{M}$ denotes the trainable memory queries and $\mathbf{H}$ denotes the output contextual features of the semantic encoder. $\mathbf{O}=[o_1,\dots,o_M]$ are the output of the memory module, in which $o_m$ is the semantic representation extracted by the $m$-th memory query.

\paragraph{Vector Quantization Module}

This is the core module of our DCMA model.
Inspired by vq-wav2vec \cite{DBLP:conf/iclr/BaevskiSA20}, we discretize the semantic memory $o_m$ to a finite set of virtual tokens via a quantizer, so that we can perform cross-modal alignment in discrete space. 
A codebook is a vocabulary that contains $V$ virtual tokens, each of which is represented by a vector $e\in \mathbb{R}^{d^{\prime}}$ like word embedding. The vector quantization module aims to select one entry from the codebook as the output.

As illustrated in Figure \ref{fig:gumbel}, given a semantic memory $o_m$, we first apply a linear layer, followed by GELU and another linear layer to map it into logits $l_{m} \in \mathbb{R}^{V}$, which is the score of each virtual token. Second, we adopt Gumbel softmax \cite{DBLP:conf/nips/MaddisonTM14,DBLP:conf/iclr/JangGP17} to choose the discrete entries in a differentiable way. The probabilities for selecting the $j$-th entry are
\begin{equation}
    p_{m,j}=\frac{\exp (l_{m,j} + n_j) / \tau}{\sum_{k=1}^V \exp (l_{m,k}+n_k) / \tau}
\end{equation}
where $n=-\log(-\log (u))$ and $u$ are sampled from the uniform distribution $\mathcal{U}(0,1)$. The $j$-th entry is chosen by $j=\operatorname{argmax}_jp_{m,j}$ during the forward pass, denoted as $\hat{o}_m=e_j$. The quantization module updates the original semantic memory $o_m$ with $\hat{o}_m$, and performs the same operation for all semantic memories to obtain the outuput code vectors $\hat{\mathbf{O}}\in \mathbb{R}^{M\times d}$. During the backward pass, the gradient of selecting one entry is estimated by the gradient of the true Gumbel softmax output.

Since a codebook contains a limited discrete space of size $V$, we increase the representation capability of the discrete space by increasing the number of codebooks. Suppose there are $G$ groups with $V$ entries, this module selects one entry from each group and concatenate them to obtain $\hat{o}_m\in \mathbb{R}^{G\times d^{\prime}}$, where we set $d^{\prime}= d/G$. The grouping operation can theoretically yield $V^G$ different outputs, which means we can increase the size of the discrete space exponentially. The codebooks are shared across semantic memories extracted by different memory queries, and are also shared across two modalities.

Let $(\mathbf{s},\mathbf{x})$ be an ASR training sample, the quantization module is expected to select the same codebook entries for the speech and the corresponding text, so that the representations of both modalities are aligned in the discrete space. First, the softmax function is applied to convert the logits of $i$-th group $l_{m,i}\in \mathbb{R}^{V}$ into distribution across codebooks.
\begin{gather}
    \hat{p}^{modal}_{m,i,j}=\frac{\exp (l^{modal}_{m,i,j} ) }{\sum_{k=1}^V \exp (l^{modal}_{m,i,k}) } , modal \in \{\mathbf{s},\mathbf{x}\}
\end{gather}
Then we treat the distribution of text as target and encourage the module to make the same choices for the corresponding speech.
\begin{equation}
    \mathcal{L}_{align}(\mathbf{s},\mathbf{x})  = \frac{1}{G} \sum_{m=1}^M \sum_{i=1}^G \sum_{j=1}^V -sg(\hat{p}^{\mathbf{x}}_{m,i,j})\log \hat{p}^{\mathbf{s}}_{m,i,j}
\end{equation}
where $M$,$G$,$V$ are the number of memory queries, codebook groups and entries respectively, and $sg(\cdot)$ means the \emph{stop gradient} operation.

\paragraph{Text Decoder}

The text decoder also follows the basic network structure of the Transformer, which takes the fixed-length code vectors $\hat{\mathbf{O}}$ as input, and generates the target translation conditioned on the discrete representations. The code vectors from the text are used in training while those from the speech are used in inference.

The decoder module is trained using parallel text data. Let $(\mathbf{x}^{\prime},\mathbf{y})$ be an MT training example, the objective function of the MT task can be calculated by cross-entropy loss as in:
\begin{equation}
    \mathcal{L}_{MT}(\mathbf{x}^{\prime},\mathbf{y}) = -\sum_{i=1}^{|\mathbf{y}|} \log p(\mathbf{y}_i|\mathbf{y}_{<i},\mathbf{x}^{\prime})
\end{equation}

\subsection{Training Process} \label{sec:training}

We train our model in the pretrain-finetune manner. We first train the semantic encoder, shared memory module, shared vector quantization module and the text decoder with the MT task and the MLM task. It helps to make the training more stable and enriches the codebook entries with semantic information. In the finetune stage, we optimize the entire model with all the relevant tasks as shown below:
\begin{align} \label{eq:loss}
    \mathcal{L} = & \mathbb{E}_{(\mathbf{s},\mathbf{x})\in \mathcal{D}_{ASR}} [\mathcal{L}_{align}(\mathbf{s},\mathbf{x}) + \alpha \mathcal{L}_{CTC}(\mathbf{s},\mathbf{x})] \\ \nonumber
    & + \mathbb{E}_{(\mathbf{x}^{\prime},\mathbf{y})\in \mathcal{D}_{MT}} [\mathcal{L}_{MT}(\mathbf{x}^{\prime},\mathbf{y}) + \beta \mathcal{L}_{MLM}(\mathbf{x}^{\prime})] 
\end{align}
We optimize $\mathcal{L}_{align}$ and $\mathcal{L}_{CTC}$ in the ASR batches, and alternately optimize the $\mathcal{L}_{MT}$ and $\mathcal{L}_{MLM}$ in the MT batches. Note that no end-to-end ST data are involved in the training process.

\section{Experiments}

\subsection{Datasets}

\paragraph{ASR Datasets}
MUST-C \citep{DBLP:conf/naacl/GangiCBNT19} is one of the largest multilingual speech translation datasets. MUST-C contains the English (En) speech, the corresponding transcription, and the target translation in 8 different languages, including German (De), French (Fr), Russian (Ru), Spanish (Es), Romanian (Ro), Italian (It), Portuguese (Pt), and Dutch (Nl). During training, we use only the speech and its transcription as ASR dataset. During inference, we use the \emph{dev} set for validation and the \emph{tst-COMMON} set for test.

\paragraph{MT Datasets} 
We use MT datasets in various domains different from the ASR dataset. Specifically, we use WMT 2014\footnote{\url{http://www.statmt.org/wmt14/translation-task.html}} for En-De, En-Fr, En-Ru and En-Es, WMT 2016\footnote{\url{https://www.statmt.org/wmt16/translation-task.html}} for En-Ro, and OPUS100\footnote{\url{http://opus.nlpl.eu/opus-100.php}} for En-It, En-Pt and En-Nl. The transcription and its translation in MUST-C can serve as in-domain MT data to further investigate the performance of zero-shot ST\footnote{Our method leverages MT batches and ASR batches alternately, so the source language text overlap brought by introducing in-domain MT data will not cause data leakage.}. The detailed statistics are shown in Table \ref{tab:data_sta}.

\begin{table}[t]
\footnotesize
\centering
\begin{threeparttable}
\begin{tabular}{l|cc|cc}
    \toprule
    \multirow{2}{*}{En$\rightarrow$} & \multicolumn{2}{c}{ASR} & \multicolumn{2}{c}{MT} \\ 
     & hours & \#sentences &  name & \#sentences \\ \midrule
    De & 408 & 234K &  WMT14 & 4.5M \\
    Fr & 492 & 280K &  WMT14 & 5.4M\tnote{*}\\
    Ru & 489 & 270K &  WMT14 & 1.0M\\
    Es & 504 & 270K &  WMT14 & 3.8M\\
    Ro & 432 & 240K &  WMT16 & 0.6M \\
    It & 465 & 211K &  OPUS100 & 1.0M \\
    Pt & 385 & 211K &  OPUS100 & 1.0M\\
    Nl & 442 & 253K &  OPUS100 & 1.0M \\
    \bottomrule
\end{tabular}
\begin{tablenotes}
    \item[*] We only use \emph{europarl v7}, \emph{commoncrawl} and \emph{news commentary} subsets of WMT14 En-Fr.
\end{tablenotes}
\caption{The detailed statistics of all datasets.}
\label{tab:data_sta}
\end{threeparttable}
\end{table}

\subsection{Experimental Settings}

\paragraph{Pre-processing}

For speech input, we use the 16 kHZ raw audio waves and normalize the wave sequences by a factor of $2^{15}$ to the range of $[-1,1]$. In order to utilize the GPU more efficiently, we filter out speech-transcription pairs whose audio frames exceed 1M.

For the text input,  capitalization and punctuation are preserved. We filter out MT samples whose number of source or target tokens is over 250 and whose length ratio is outside the $[2/3,3/2]$ interval. For each language pair, we use a  unigram sentencepiece\footnote{\url{https://github.com/google/sentencepiece}} model to learn a 10K vocabulary from the text portion of MUST-C, and apply it to segment other text data into subword units. The vocabulary is shared across both source and target languages.

\begin{table*}[htbp]
\footnotesize
\centering
\resizebox{\textwidth}{!}{
\begin{tabular}{lcccccccccccc}
    \toprule
    \multirow{2}{*}{Methods} & \multicolumn{4}{c}{Training Data} & \multicolumn{8}{c}{BLEU} \\ \cmidrule(lr){2-5} \cmidrule(lr){6-13}
     & Speech & ASR & MT & ST & En-De & En-Fr & En-Ru & En-Es & En-Ro & En-It & En-Pt & En-Nl \\ \midrule
    \multicolumn{13}{c}{\emph{Previous state-of-the-art for zero-shot ST}}\\ \midrule
    MultiSLT \citep{DBLP:conf/asru/EscolanoCFS21} & $\times$ & \checkmark & \checkmark & $\times$ & 6.8 & 10.9 & - & 6.8 & - & - & - & -\\ \midrule
    \multicolumn{13}{c}{\emph{Previous cross-modal alignment methods}}\\ \midrule
    Chimera* \cite{DBLP:conf/acl/HanWJL21} & \checkmark & \checkmark & \checkmark & $\times$ & 13.5 & 22.2 & 8.3 & 15.3 & 8.5 & 12.6 & 16.9 & 13.1\\ \midrule
    \multicolumn{13}{c}{\emph{Supervised baselines on MUST-C}}\\ \midrule
    Fairseq ST \cite{DBLP:conf/ijcnlp/WangTMWOP20} & $\times$ & \checkmark & $\times$ & \checkmark & 22.7 & 32.9 & 15.3 & 27.2 & 21.9 & 22.7 & 28.1 & 27.3\\ 
    Espnet ST \cite{DBLP:conf/acl/InagumaKDKYHW20} & $\times$ & \checkmark & \checkmark & \checkmark & 22.9 & 32.8 & 15.8 & 28.0 & 21.9 & 23.8 & 28.0 & 27.4 \\
    W2V2-Transformer** & \checkmark & $\times$ & \checkmark & \checkmark & 24.1 & 35.0 & 16.3 & 29.4 & 23.1 & 24.8 & 30.0 & 28.9 \\ \midrule
    \multicolumn{13}{c}{\emph{This work}}\\ \midrule
    DCMA & \checkmark & \checkmark & \checkmark & $\times$ & 22.4 & 29.7 & 11.8 & 24.6 & 16.8 & 18.4 & 24.2 & 22.0 \\
    \quad + in-domain MT data & \checkmark & \checkmark & \checkmark & $\times$ & 24.0 & 33.1 & 16.0 & 26.2 & 22.2 & 24.1 & 29.2 & 28.3\\
    \bottomrule
\end{tabular}
}
\caption{BLEU scores on the tst-COMMON set in 8 language pairs in MUST-C. ``Speech'' means speech self-supervised pretraining using unlabeled audio data. ASR data is leveraged for speech recognition task or for cross-modal alignment. * is reproduced under zero-shot scenario, which is a strong baseline of performing cross-modal alignment in continuous space. ** from \citet{DBLP:journals/corr/abs-2203-10426} is a baseline model by combining wav2vec 2.0 \cite{DBLP:conf/nips/BaevskiZMA20} and a Transformer.}
\label{tab:main_res}
\end{table*}

\paragraph{Model Configuration}

We use wav2vec2.0 \cite{DBLP:conf/nips/BaevskiZMA20} as the acoustic encoder, which follows the base configurations and is pretrained on the unlabeled audio data from LibriSpeech \citep{DBLP:conf/icassp/PanayotovCPK15}. Two additional 1-dimensional convolution layers are used to shrink the length of the speech features, with stride size 2, kernel size 5, padding 2, and hidden dimension 1024. 

For the semantic encoder, we use a 6-layer Transformer encoder. The memory queries are 64 512-dimensional vectors. The vector quantization module consists of $G=128$ groups of codebook with $V=50$ entries in each group, which can produce $50^{128}$  possible codewords. A linear layer, followed by GELU and another linear layer are used to project the semantic memory into $G\cdot V=6400$ logits with 1024 hidden units.  The Gumbel softmax produces a one-hot vector for each group. The temperature $\tau$ decays exponentially from 2 to 0.5 with a factor of 0.999995 and then keeps constant at 0.5. The text decoder consists of 6 transformer layers. Each of the layers in the semantic encoder and text decoder module has 512-dimensional hidden sizes, 8 attention heads, and 2048 feed-forward hidden units. A 512-dimensional word embedding layer is shared across the semantic encoder and the text decoder.

\paragraph{Training Details}

We train our model following the pretrain-finetune strategy. During pretraining, we train the model with the MT and MLM tasks. The learning rate is 7e-4 with 4K warm-up updates. We pretrain the model up to 150K updates, with at most 1152 sentence pairs per batch. In the stage of zero-shot finetune, we adopt multi-task training as described in Section \ref{sec:training}. We set both $\alpha$ and $\beta$ in Equation (\ref{eq:loss}) to 1.0. The learning rate is set to 1e-4 with 10K warm-up updates. We finetune the model up to 150K updates, with at most 16M audio frames per batch. In both the pretrain and the finetune stages, the model is trained by Adam optimizer \citep{DBLP:journals/corr/KingmaB14} with $\beta_1=0.9,\beta_2=0.98$. An inverse square root schedule algorithm is adopted for the learning rate. Both the dropout and label smoothing rate are set to 0.1 for regularization. The whole training process is carried out on two Nvidia Tesla-V100 GPUs.

During inference, we average the model parameters of the last 5 checkpoints. We use beam search with a beam size of 5. The performance is evaluated with case-sensitive BLEU \citep{DBLP:conf/acl/PapineniRWZ02} calculated by SacreBLEU\footnote{\url{https://github.com/mjpost/sacrebleu}} \citep{DBLP:conf/wmt/Post18}.

\section{Results and Analysis}

\subsection{Main Results}

\paragraph{Comparison with zero-shot methods}

We compare our DCMA method with MultiSLT \citep{DBLP:conf/asru/EscolanoCFS21}, which is the previous SOTA for zero-shot speech translation. As shown in Table \ref{tab:main_res}, our method achieves remarkable improvements. We also notice that introducing in-domain MT data can further improve the performance and is more useful when the MT data size is small (Ru, Ro, It, Pt and Nl)\footnote{This can be further demonstrated in Appendix \ref{app:mt}}. 
To demonstrate the advantages of discrete alignment, we implement the Chimera \citep{DBLP:conf/acl/HanWJL21}, a continuous alignment method, to perform zero-shot ST, which has a similar model architecture as our DCMA and is trained under the same conditions. The difference is that it does not have the vector quatization module and instead aligns the representations of speech and text with the continuous contrastive loss. Our method significantly outperforms Chimera on all language pairs, demonstrating the potential of discrete space for cross-modal alignment in zero-shot ST\footnote{We conduct some analyses of the representations learnt by continuous alignment and discrete alignment in Appendix \ref{app:vs}}.

\paragraph{Comparison with supervised methods}

We compare our zero-shot DCMA method with the supervised baselines Fairseq ST \citep{DBLP:conf/ijcnlp/WangTMWOP20}, Espnet ST \citep{DBLP:conf/acl/InagumaKDKYHW20}, and W2V2-Transformer, which also adopt the pretrain-finetune procedure but are finetuned on the parallel ST data. W2V2-Transformer utilizes a speech self-supervised learning model, combining the pretrained wav2vec 2.0, a 6-layers transforemr encoder and a 6-layers transforemr decoder. As shown in Table \ref{tab:main_res}, our  DCMA method achieves competitive performance compared to the supervised baselines when in-domain MT data are used. This demonstrates that the zero-shot DCMA method can learn an effective end-to-end ST model without using end-to-end ST data, as well as the possibility of projecting speech and text into a shared space. 
Although our zero-shot method uses more MT data than the supervised baselines, our method does not use any end-to-end ST data, making it widely useful in low-resource scenarios.

\paragraph{Comparison with cascaded system and data synthesis method}

We compare our zero-shot DCMA method with those that also do not use parallel ST data, namely the cascade system and data synthetic method \citep{DBLP:conf/icassp/JiaJMWCCALW19}. 
For the cascade ST system, the ASR part is the W2V2-Transformer, and the MT part follows the basic Transformer configuration. The cascaded system first translates the speech into source language text, and then translates the transcription into the target translation. 
For generating synthetic data, we first leverage the MT model in cascaded system to translate the transcriptions in ASR dataset into the target translation. The ST model W2V2-Transformer is initialized with the MT model and finetuned with the synthetic data.  As shown in Table \ref{tab:indomain}, our DCMA method achieves comparable performance to those of other methods. 
However, cascaded system faces the problem of high decoding latency, and generating synthtic data is a time-consuming process. 
In Section \ref{sec:acc}, we show that our method can outperform the cascade system on well-aligned subsets.

\begin{table}[t]
    \centering
    \small
    \resizebox{\linewidth}{!}{
    \begin{tabular}{l|ccc}
    \toprule
    Methods & WER($\downarrow$)  & MT BLEU & ST BLEU \\
    \midrule
    Cascaded ST & 11.1 & 28.6  &  23.5 \\ 
    \quad + in-domain MT data & 11.1 & 32.4 & 26.7 \\
    Synthetic Data & - & 28.6 & 23.3 \\
    DCMA & - & - & 22.4 \\
    \quad + in-domain MT data & - & - & 24.0 \\
    \bottomrule
    \end{tabular}
    }
    \caption{Comparison with zero-resource methods on MUST-C En-De corpus. We report the Word Error Rate (WER) of speech-transcription pairs for ASR models, the MT BLEU scores of transcription-translation pairs for MT models, and the ST BLEU scores of speech-translation pairs for ST models.}
    \label{tab:indomain}
\end{table}

\begin{table}[t]
    \small
    \centering
    \begin{tabular}{cc|c}
    \toprule
    Parameter Sharing & Discrete Alignment & BLEU \\
    \midrule
    \checkmark  & \checkmark & 22.4 \\
    \checkmark  &  $\times$ & 1.3 \\
    $\times$  & \checkmark  & 21.8 \\
    $\times$  & $\times$  & 0.1 \\
    \bottomrule
    \end{tabular}
    \caption{Ablation studies on the MUST-C En-De corpus. ``Parameter sharing'' means sharing the softmax vocabulary and parameters across MLM and CTC.}
    \label{tab:ablation}
\end{table}

\subsection{Ablation Studies}

We share the softmax vocabulary and parameters across the two training objectives, $\mathcal{L}_{MLM}$ and $\mathcal{L}_{CTC}$, to encourage implicit alignment between the speech and text representations learnt by the semantic encoder. To better evaluate the contribution of the sharing strategy and our proposed discrete alignment model, we conduct ablation studies on the MUST-C En-De corpus. As shown in Table \ref{tab:ablation}, implicit alignment between speech and text through the sharing strategy is beneficial to improve the performance. However, the proposed discrete alignment method is the most important and indispensable (performance degradation from 22.4 to 1.3 without it).

\begin{table}[t]
    \centering
    \small
    \begin{tabular}{l|cc|cc}
    \toprule
    Methods & V & G & MT BLEU & ST BLEU  \\
    \midrule
    Transformer & - & - & 28.6 & -\\
   % Chimera & - & - & 28.7 & -\\
    DCMA &  50 & 2 & 4.5 & -\\
    DCMA &  50 & 4 & 19.6 & -\\
    DCMA &  50 & 8 & 25.0 & -\\
    DCMA &  50 & 16 & 25.7 & -\\
    DCMA &  50 & 32 & 26.7 & 18.5 \\
    DCMA &  50 & 64 & 27.4 & 21.5 \\
    DCMA &  50 & 128 & 28.0 & 22.4 \\
    DCMA &  50 & 256 & 27.8 & 19.8 \\
    \bottomrule
    \end{tabular}
    \caption{MT BLEU scores and ST BLEU scores on the tst-COMMON set of MUST-C En-De corpus with different size of codebooks. The number of theoretically possible outputs is $V^G$.}
    \label{tab:codebook}
\end{table}

\subsection{Effect of the Size of Codebooks}

The shared vector quantization module discretizes the continuous vectors to a finite set of virtual tokens, so we can perform cross-modal alignment in the shared discrete space. An important question arises that how big codebooks are needed so that the vectors can be discretized without losing representation ability. Given $G$ groups of codebook with $V$ entries, the number of theoretically possible outputs is $V^G$, so that we can exponentially increase the size of codebooks by increasing $G$. We vary the setting of $G$ and report the MT BLEU scores of transcription-translation pairs and the zero-shot ST BLEU scores of speech-translation pairs. We observe that when the discrete space is small (e.g. row 2), the quantization operation loses a great deal of representational power, but when the discrete space becomes larger the MT performance gets better and better. However, continuing to increase the size of codebooks (e.g. when G is 256) does not improve the performance. Our proposed DCMA method achieves the best performance when the number of groups is set to $G=128$.

\subsection{Effect of the Size of ASR Data}

\begin{figure}[t]
    \centering
    \includegraphics[width=\linewidth]{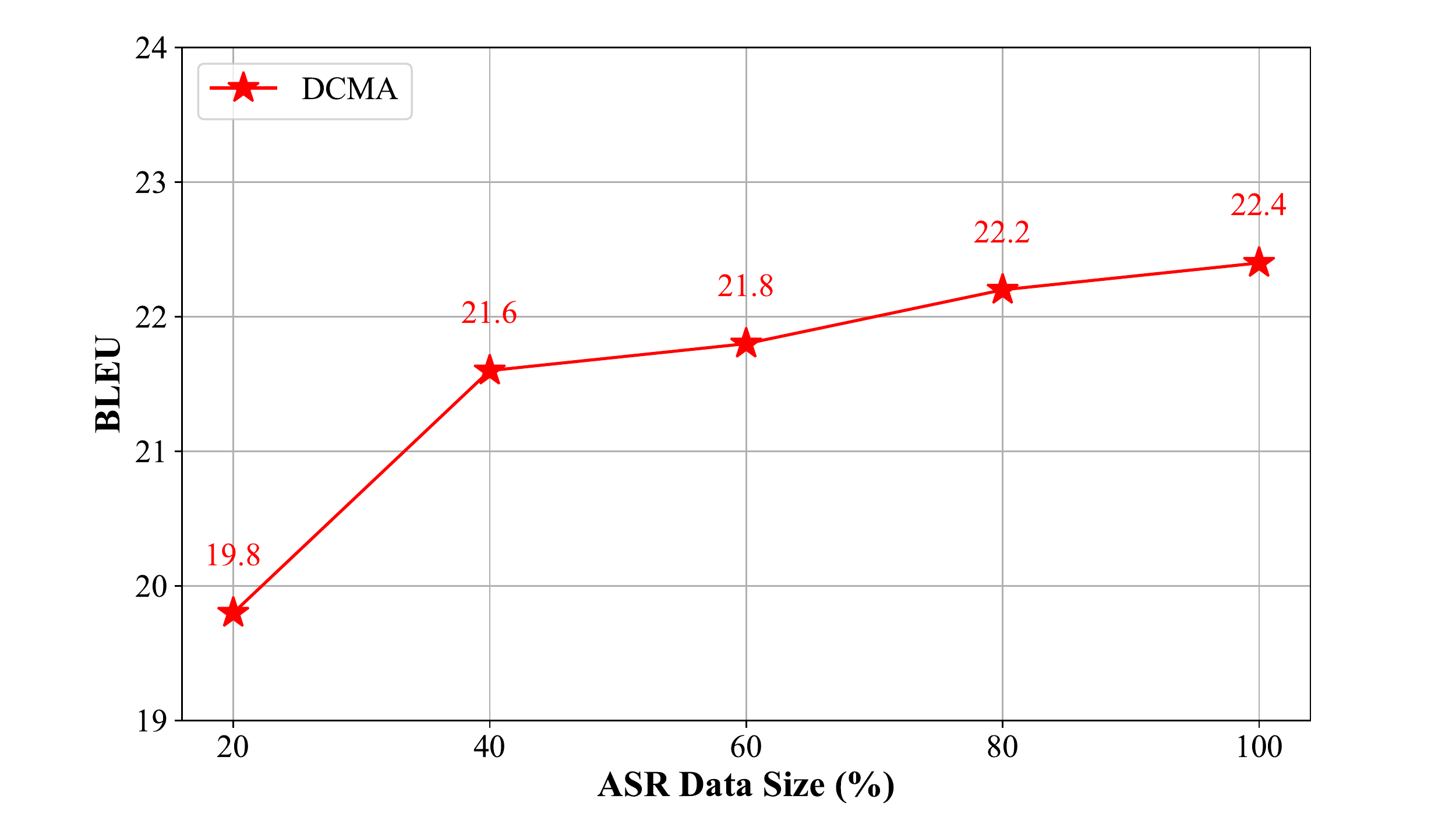}
    \caption{Curve of BLEU scores against the size of ASR data size on MUST-C En-De corpus.}
    \label{fig:asr}
\end{figure}

The key part of our method is to use ASR data to learn a discrete shared semantic space. Therefore, the ASR data size is an important factor. We randomly sample different amount of ASR data from the MUST-C En-De corpus. As shown in Figure \ref{fig:asr}, we observe a continuous improvement of BLEU scores with the increase of ASR data size.

\begin{figure}[t]
    \centering
    \includegraphics[width=\linewidth]{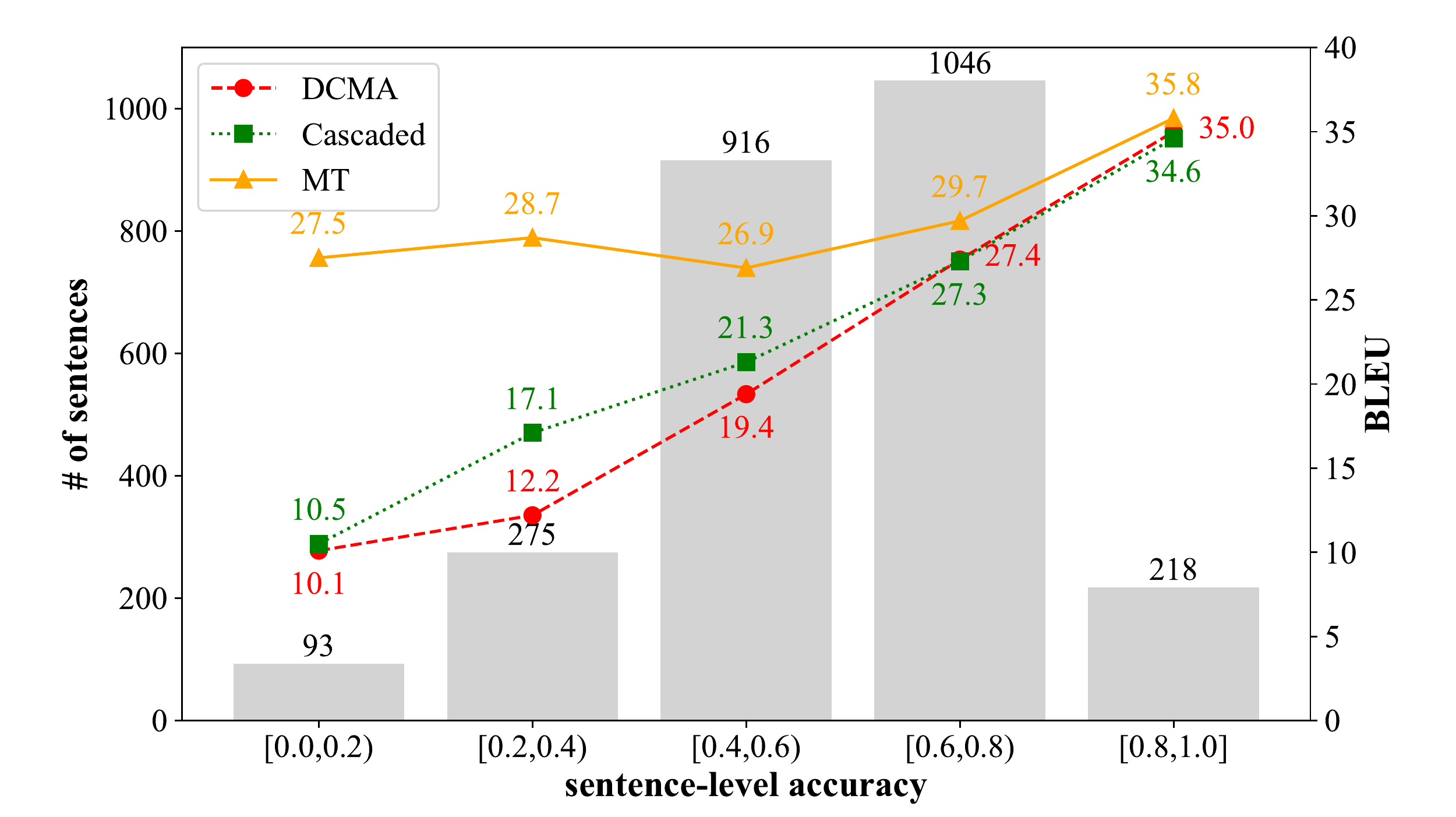}
    \caption{The tst-COMMON set of MUST-C En-De corpus is divided into 5 subsets according to sentence-level alignment accuracy. The histogram represents the size of each subset. Red circles are the ST BLEU scores of DCMA, green squares are the ST BLEU scores of cascaded system, and orange triangles are the transcription-translation BLEU scores of MT model. Our method is comparable to cascaded system and text translation on the well-aligned subsets.}
    \label{fig:sentence}
\end{figure}

\subsection{Can Our Method Achieve Cross-modal Alignment?} \label{sec:acc}

To evaluate whether our model can project the speech and text with the same semantic to the same virtual tokens, we conduct some analyses of the alignment accuracy. Let $(\mathbf{s},\mathbf{x},\mathbf{y})$ be an ST test sample. The speech input $\mathbf{s}$ can be discretized into $\mathbf{Z^s}=[\mathbf{z^s_1},\dots,\mathbf{z^s_M}]$, where $\mathbf{z^{\mathbf{s}}_i}=[z^{\mathbf{s}}_{i1},\dots,z^{\mathbf{s}}_{iG}]$ is the code vector ids selected when discretizing the features extracted by the $i$-th memory query and $z^{\mathbf{s}}_{ij}$ is the code vector id selected in the $j$-th group. The $M$ and $G$ are the number of memory queries and the number of codebook groups respectively. We do the same operation for text input $\mathbf{x}$ to obtain $\mathbf{Z^x}$, and define the sentence-level accuracy $sent\_acc = \frac{\sum_{i=1}^M\sum_{j=1}^G \mathbf{1}\{z^{\mathbf{s}}_{ij}=z^{\mathbf{x}}_{ij}\}}{M\cdot G}$. The test set is divided into 5 subsets according to the sentence-level alignment accuracy, and we calculate the MT BLEU scores and ST BLEU scores for each subset. As shown in Figure \ref{fig:sentence}, most of speech utterances are discretized with over 40\% alignment accuracy, which indicates the ability of our model to align speech and text into shared discrete codebooks. We also observe a continuous improvement of ST BLEU scores with the increase of sentence-level alignment accuracy. 
The performance of the zero-shot ST is comparable to that of text translation or better than that of the cascade system on the well-aligned subsets. It indicates the big potential of our method that the zero-shot ST will achieve much better performance if we can design better cross-modal alignment method.

\section{Conclusion}

In this paper, we propose a novel alignment method DCMA to enable zero-shot ST. The key part of our approach is to discretize the continuous vectors to a finite set of virtual tokens and use ASR data to map the corresponding speech and text to the same virtual token in the shared codebook. Our experiments demonstrate that our method can learn an effective end-to-end ST model without any parallel ST data. It significantly improves the existing SOTA and achieves competitive performance compared to the supervised models.

\section{Limitations}

In this paper, we propose a zero-shot ST method, which eliminates reliance on end-to-end ST data, allowing end-to-end models to be trained on the same data conditions as cascade systems. The performance of the cascade systems can benefit from both the pretrained speech models (such as wav2vec 2.0) and the pretrained text models (such as BART and T5). However, since our method introduces additional modules between the encoder and the decoder, the pretrained text model cannot be directly integrated into the architecture. Experiments show that our method can outperform the cascade system and obtain comparable results to those of text translation on the well-aligned subsets. However, on the examples with low alignment accuracy, our method is not as robust as the cascade system. How to improve projections onto discrete units is an issue that our future work will explore.

\section*{Acknowledgements}

This work is supported by the Natural Science Foundation of China under Grant 62122088 and U1836221.

\bibliography{anthology,custom}

\appendix
\section{MT BLEU Scores after Pretraining}
\label{app:mt}

To evaluate the domain gap between MUST-C and the MT datasets (WMT and OPUS), we report MT BLEU scores after pretraining. 
% As shown in Table \ref{tab:mt}, introducing in-domain MT data significantly improves the MT performance, especially when the MT data size is small (Ru, Ro, It, Pt and Nl). 
% This also proves the necessity of introducing in-domain MT data to further investigate the performance of zero-shot ST.
\begin{table}[htbp]
\small
\centering
\begin{tabular}{l|ccc}
    \toprule
    En$\rightarrow$ & DCMA & + in-domain MT data & $\Delta$ \\ \midrule
    De & 28.0 & 31.2 & +3.2 \\
    Fr & 39.3 & 43.2 & +3.9\\
    Ru & 14.5 & 19.6 & +5.1\\
    Es & 31.6 & 35.7 & +4.1\\
    Ro & 21.1 & 29.1 & +8.0 \\
    It & 23.3 & 30.8 & +7.5\\
    Pt & 29.5 & 36.6 & +7.1\\
    Nl & 27.3 & 35.0 & +7.7\\
    \bottomrule
\end{tabular}
\caption{MT BLEU scores on transcription-translation pairs of MUST-C tst-COMMON set.}
\label{tab:mt}
\end{table}

\section{Discrete vs. Continuous Alignment}
\label{app:vs}

To explore the benefits of discrete alignment, we conduct some analyses of representations learnt by continuous alignment and discrete alignment. Let $(\mathbf{s_i},\mathbf{x_i})$ be an ASR test sample. The fixed-length representations produced by the encoder are denoted as $\mathbf{O^s_i}=[o^s_{i1},\dots,o^s_{iM}]$ and $\mathbf{O^x_i}=[o^x_{i1},\dots,o^x_{iM}]$, where $M$ is the number of memory queries. We define the sentence embedding in each modality $\mathbf{\bar{O}^s_i} = \frac{1}{M} \sum_{j=1}^M{o^s_{ij}}$, $\mathbf{\bar{O}^x_i} = \frac{1}{M} \sum_{j=1}^M{o^x_{ij}}$.
Then we calculate the average memory-level and sentence-level cosine similarity on subsets with different alignment accuracy, as described in Section \ref{sec:acc}.
\begin{gather*}
    sim\_memory = \frac{1}{N \cdot M} \sum_{i=1}^N \sum_{j=1}^M{\cos (o^s_{ij},o^x_{ij})} \\
    sim\_sentence = \frac{1}{N} \sum_{i=1}^N{\cos(\mathbf{\bar{O}^s_i},\mathbf{\bar{O}^x_i})}
\end{gather*}
As shown in Table \ref{tab:sim}, our discrete alignment method significantly improves the sentence-level cosine similarity over the continuous alignment, though both alignments are performed in memory level. 
We believe it is because that the discrete alignment aligns the corresponding speech and text semantically, rather than just minimizing the distance gap between memories. 
Our method can also get better memory-level cosine similarity on the well-aligned subsets.

\begin{table}[h]
\footnotesize
\centering
\begin{tabular}{l|cc|cc}
    \toprule
    \multirow{2}{*}{Acc} & \multicolumn{2}{c}{DCMA} & \multicolumn{2}{c}{Chimera} \\ 
     & memory & sentence &  memory & sentence \\ \midrule
    $\left[ 0.8, 1.0\right]$ & 0.92 & 0.94 &  0.87 & 0.70 \\
    $\left[ 0.6, 0.8\right)$ & 0.84 & 0.89 &  0.81 & 0.63 \\
    $\left[ 0.4, 0.6\right)$ & 0.69 & 0.80 &  0.71 & 0.56 \\
    $\left[ 0.2, 0.4\right)$ & 0.48 & 0.67 &  0.58 & 0.47 \\
    $\left[ 0.0, 0.2\right)$ & 0.28 & 0.57 &  0.38 & 0.36 \\\midrule
    $\left[ 0.0, 1.0\right]$ & 0.73 & 0.82 & 0.74 & 0.58 \\
    \bottomrule
\end{tabular}
\caption{Comparison of memory-level and sentence-level representation similarity on different subsets.}
\label{tab:sim}
\end{table}

% This is an appendix.

\end{document}